\documentclass[letterpaper, 10 pt, conference]{ieeeconf}  

\IEEEoverridecommandlockouts                              

\usepackage{xspace}
\usepackage{epsfig}
\usepackage{graphicx}
\usepackage{amsmath}
\usepackage{amssymb}
\usepackage[noadjust]{cite}
\usepackage{color, colortbl}
\usepackage[pagebackref=true,breaklinks=true,bookmarks=false]{hyperref}
\usepackage{array}

\newcommand{\bc}{\mathbf{c}}

\newcommand{\bn}{\mathbf{n}}

\newcommand{\bs}{\mathbf{s}}

\newcommand{\bx}{\mathbf{x}}

\newcommand{\nR}{\mathbb{R}}

\newcommand{\cC}{\mathcal{C}}

\newcommand{\cL}{\mathcal{L}}

\newcommand{\cN}{\mathcal{N}}

\newcommand{\cS}{\mathcal{S}}

\newcommand{\tabref}[1]{Table~\ref{#1}}

\makeatletter
\DeclareRobustCommand\onedot{\futurelet\@let@token\@onedot}
\def\@onedot{\ifx\@let@token.\else.\null\fi\xspace}
\def\eg{e.g\onedot} 
\def\ie{i.e\onedot}

\makeatother

\newcommand{\boldparagraph}[1]{\vspace{0.1cm}\noindent{\bf #1:}}
\newcommand{\boldquestion}[1]{\vspace{0.1cm}\noindent{\bf #1}}

\definecolor{darkgreen}{rgb}{0,0.7,0}
\definecolor{brown}{rgb}{0.59,0.29,0.00}
\definecolor{gray}{gray}{0.9}

\title{\LARGE \bf
Label Efficient Visual Abstractions for Autonomous Driving
}

\author{Aseem Behl*$^{1,2}$, Kashyap Chitta*$^{1,2}$, Aditya Prakash$^{1}$, Eshed Ohn-Bar$^{1,3}$ and Andreas Geiger$^{1,2}$
\thanks{* indicates equal contribution, listed in alphabetical order. $^{1}$Max Planck Institute for Intelligent Systems, T\"ubingen; $^{2}$University of T\"ubingen; $^{3}$Boston University. {\tt\small \{firstname.lastname\}@tue.mpg.de}}%
}

\begin{document}

\maketitle
\thispagestyle{empty}
\pagestyle{empty}

\maketitle

\begin{abstract}

It is well known that semantic segmentation can be used as an effective intermediate representation for learning driving policies. However, the task of street scene semantic segmentation requires expensive annotations. Furthermore, segmentation algorithms are often trained irrespective of the actual driving task, using auxiliary image-space loss functions which are not guaranteed to maximize driving metrics such as safety or distance traveled per intervention. In this work, we seek to quantify the impact of reducing segmentation annotation costs on learned behavior cloning agents. We analyze several segmentation-based intermediate representations. We use these \textit{visual abstractions} to systematically study the trade-off between annotation efficiency and driving performance, \ie, the types of classes labeled, the number of image samples used to learn the visual abstraction model, and their granularity (\eg, object masks vs. 2D bounding boxes). Our analysis uncovers several practical insights into how segmentation-based visual abstractions can be exploited in a more label efficient manner. Surprisingly, we find that state-of-the-art driving performance can be achieved with orders of magnitude reduction in annotation cost. Beyond label efficiency, we find several additional training benefits when leveraging visual abstractions, such as a significant reduction in the variance of the learned policy when compared to state-of-the-art end-to-end driving models.

\end{abstract}

\section{Introduction}
Significant research effort has been devoted into semantic segmentation of street scenes in recent years, where images from a camera sensor mounted on a vehicle are segmented into classes such as road, sidewalk, and pedestrian~\cite{Brostow2009PRL,Cordts2016CVPR,Yu2018ARXIV,Huang2018ARXIV,Geiger2012CVPR}. It is widely believed that this kind of accurate scene understanding is key for robust self-driving vehicles. Existing state-of-the-art methods~\cite{Zhao2017CVPR} optimize for image-level metrics such as mIoU, which is challenging as it requires a combination of coarse contextual reasoning and fine pixel-level accuracy \cite{Farabet2013PAMI}. The emphasis on such image-level requirements has resulted in large segmentation benchmarks, \ie, thousands of images, with high labeling costs. However, the development of such benchmarks, in terms of annotation type and cost, is often done independently of the actual \textit{driving task} which encompasses optimizing metrics such as distance traveled per human intervention.

\begin{figure}[t!]
	\centering
	\small
	\begin{tabular}{c}
		\includegraphics[width=0.95\columnwidth]{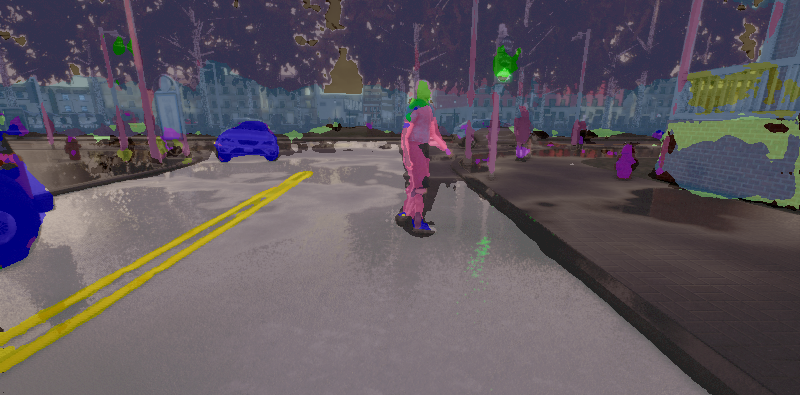}\\
		\rowcolor{gray}
		Trained with 6400 finely annotated images and 14 classes\\
		\rowcolor{gray}
		\textbf{Annotation time $\approx$ 7500 hours, policy success rate = 50\%}\\
		\\
		\includegraphics[width=0.95\columnwidth]{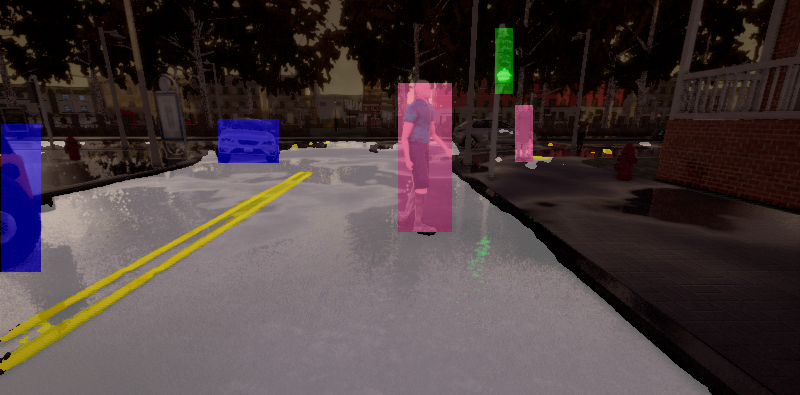}\\
		\rowcolor{gray}
		Trained with 1600 coarsely annotated images and 6 classes\\
		\rowcolor{gray}
		\textbf{Annotation time $\approx$ 50 hours, policy success rate = 58\%}\\
	\end{tabular}
	\caption{
	\textbf{Label efficient visual abstractions for learning driving policies.} To address issues with obtaining time-consuming annotations, we analyze image-based representations that are both \textit{\textbf{efficient}} in terms of annotation cost (\eg, bounding boxes), and \textit{\textbf{effective}} when used as intermediate representations for learning a robust driving policy. Considering six coarse safety-critical semantic categories and combining non-salient classes (\eg, sidewalk and building) into a single class can significantly reduce annotation cost while at the same time resulting in more robust driving performance. 
	}
	\label{fig:teaser}
	\vspace{-0.4cm}
\end{figure}

In parallel, there has been a surge in interest on using \textit{visual priors} for learning end-to-end control policies with improved performance, generalization and sample efficiency~\cite{Zhou2019SR,Mousavian2019ICRA,Sax2019CORL}. Instead of learning to act directly from image observations, which is challenging due to the high-dimensional input, a visual prior is enforced by decomposing the task into intermediate sub-tasks. These intermediate visual sub-tasks, \eg, object detection, segmentation, depth and motion estimation, optimized independently, are then fed as an input to a policy learning algorithm, \eg,~\cite{Sax2019CORL}. In particular, semantic segmentation has been shown to act as a powerful visual prior for driving agents \cite{Shwartz2016ARXIV,Muller2018CORL}. While beneficial for learning robust policies, such intermediate sub-tasks require explicit supervision in the form of additional manual annotations. For several visual priors, obtaining these annotations can be time consuming, tedious, and prohibitive. 

A useful visual prior needs to encode the right assumptions about the environment in order to simplify policy learning. In the case of autonomous driving, semantic segmentation encodes the fact that certain pixels in an image can be treated similarly: \eg the agent can drive on roads but not sidewalks; the agent must not collide with other vehicles or pedestrians. However, it is unclear which semantic classes are relevant to the driving task and to which granularity they should be labeled. This motivates our study of \textit{visual abstractions}, which are compact semantic segmentation-based representations of the scene with fewer classes, coarser annotation, and learned with little supervision (only several hundred images). We consider the following question:
{when used as visual priors for policy learning, are representations obtained from datasets with lower annotation costs competitive in terms of driving ability?}

Towards addressing this research question, we systematically analyze the performance of varying intermediate representations on the recent NoCrash benchmark of the CARLA urban driving simulator~\cite{Dosovitskiy2017CORL,Codevilla2019ICCV}. Our analysis uncovers several new results regarding label efficient representations. Surprisingly, we find that certain visual abstractions learned with only a fraction of the original labeling cost can still perform as well or better when used as inputs for training behavior cloning policies (see Fig.~\ref{fig:teaser}). Overall, our contributions are three-fold:
\begin{itemize}
    \item Given the same amount of training data, we empirically show that using classes less relevant to the driving policy can lead to degraded performance. We find that only few of the commonly used classes are directly relevant for the driving task.
    \item We demonstrate that despite requiring only a few hundred annotated images in addition to the expert driving demonstrations, training a behavior cloning policy with visual abstractions can significantly outperform methods which learn to drive from raw images, as well as existing state-of-the-art methods that require a prohibitive amount of supervision.
    \item We further show that our visual abstractions lead to a large variance reduction when varying the training seed which has been identified as a challenging problem in imitation learning~\cite{Codevilla2019ICCV}.
\end{itemize}

Our code is available at \url{https://github.com/autonomousvision/visual_abstractions}
\section{Related Work}
This work relates to visual priors for robotic control and behavior cloning methods for autonomous driving. In this section, we briefly review the most related works.

\boldparagraph{Semantic Segmentation} Segmentation of street scenes has received increased interest in robotics and computer vision due to its implications for autonomous vehicles, with several benchmarks and approaches released in recent years \cite{Cordts2016CVPR,Yu2018ARXIV,Huang2018ARXIV}. Progress has been achieved primarily through methods that use supervised learning, with architectural innovations that improve both contextual reasoning and fine pixel-level details \cite{Long2015CVPR,Yu2017CVPR,Zhao2017CVPR}. However, generating high-quality ground truth to build semantic segmentation benchmarks is a time-consuming and expensive task. For instance, labeling a single image was reported to take 90 minutes on average for the Cityscapes dataset \cite{Cordts2016CVPR}, and approximately 60 minutes for the CamVid dataset \cite{Brostow2009PRL}. Our work focuses on reducing demands for annotation quality and quantity, which is important in the context of reducing annotation costs for segmentation and autonomous driving.

\boldparagraph{Behavior Cloning for Autonomous Driving} Behavior cloning approaches learn to map sensor observations to desired driving behavior through supervised learning. Behavior cloning for driving has historical roots~\cite{Pomerleau1988NIPS} as well as recent successes~\cite{Bojarski2016ARXIV,Zeng2019CVPR,Prakash2020CVPR,Ohn-Bar2020CVPR}. Bojarski et al.~\cite{Bojarski2016ARXIV} propose an end-to-end CNN for lane following that maps images from the front facing camera of a car to steering angles, given expert data. Conditional Imitation Learning (CIL) extends this framework by incorporating high-level navigational commands into the decision making process~\cite{Codevilla2018ICRA}. Codevilla et al.~\cite{Codevilla2019ICCV} present an analysis of several limitations of CIL. In particular, they observe that driving performance drops significantly in new environments and weather conditions. They also observe drastic variance in performance caused by model initialization and data sampling during training. The goal of this work is to address these issues with semantic input representations while maintaining low labeling costs.

\boldparagraph{Visual Priors for Improving Generalization} Recent papers have shown the effectiveness of using mid-level visual priors to improve the generalization of visuomotor policies~\cite{Wang2019ICRA,Zhou2019SR,Sax2019CORL,Mousavian2019ICRA,Muller2018CORL}. Object detection, semantic segmentation and instance segmentation have been shown to help significantly for generalization in navigation tasks~\cite{Wang2019ICRA, Zhou2019SR}. M\"{u}ller et al.~\cite{Muller2018CORL} train a policy in the CARLA simulator with a binary road segmentation as the perception input, demonstrating that learning a policy independent of the perception and low-level control eases the transfer of learned lane-keeping behavior for empty roads from simulation to a real toy car. More recently, Zhao et al.~\cite{Zhao2019ARXIV} and Toromanoff et al.~\cite{Toromanoff2020CVPR} show how to effectively incorporate knowledge from segmentation labels into a behavior cloning and reinforcement learning network, respectively. These existing studies either compare different visual priors or focus on improving policies by choosing a specific visual prior, regardless of the annotation costs. Nonetheless, knowing these costs is extremely valuable from a practitioner's perspective. In this work, we are interested in identifying and evaluating {label efficient} representations in terms of the performance and variance of the learned policies.

\boldparagraph{Compact Representations for Driving} Instead of using a comparably higher-dimensional visual prior such as pixel-level segmentation, Chen et al.~\cite{Chen2015ICCVa} present an approach which estimates a small number of human interpretable, pre-defined affordance measures such as the angle of the car relative to the road and the distance to other cars. These predicted affordances are then mapped to actions using a rule-based controller, enabling autonomous driving in the TORCS racing car simulator~\cite{Wymann2015}. Similarly, Sauer et al.~\cite{Sauer2018CORL} estimate several affordances from sensor inputs to drive in the CARLA simulator. In contrast to \cite{Chen2015ICCVa}, they consider the more challenging scenario of urban driving where the agent needs to avoid collision with obstacles on the road and navigate junctions with multiple possible driving directions. They achieve this by expanding the set of affordances to be more applicable to urban driving. These methods are relevant to our study, in that they simplify perception through compact representations. However, these affordances are hand-engineered and very low-dimensional. Thus, failures in design will lead to errors that cannot be recovered from.

\section{Method}

\begin{figure}[t]
	\centering
	\includegraphics[width=\columnwidth]{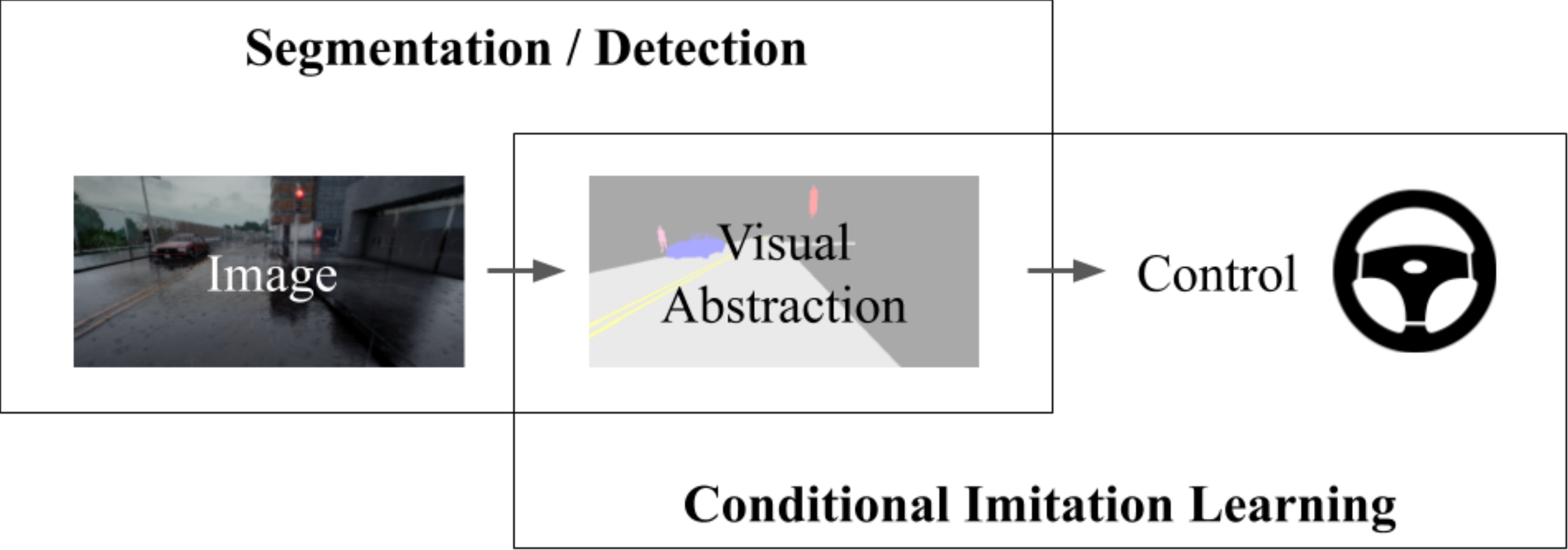}
	\caption[]{\textbf{High-level overview of the proposed study.} We investigate different segmentation-based visual abstractions by pairing them with a conditional imitation learning framework for autonomous driving.}
	\label{fig:method}
	\vspace{-0.4cm}
\end{figure}

As illustrated in Fig. \ref{fig:method}, we consider a modular approach that comprises two learned mappings, one from the RGB image to a semantic label map and one from the semantic label map to control. To learn these mappings, we use two image-based datasets, (i) $S=\{\bx^i,\bs^i\}_{i=1}^{n_s}$ which consists of $n_s$ images annotated with semantic labels, and (ii) $C=\{\bx^i,\bc^i\}_{i=1}^{n_c}$ which consists of $n_c$ images annotated with expert driving controls. First, we train the parameters of a visual abstraction model $a_\phi$ parameterized by $\phi$ using the segmentation dataset $S$. The trained visual abstraction stack is then applied to transform $C$ resulting in a control dataset $C_\phi=\{a_\phi(\bx^i),\bc^i\}_{i=1}^{n_c}$ on which we train a driving policy $\pi_\theta$ with parameters $\theta$. At test time, control values are obtained for an image $\bx^*$ by composing the two learned mappings, $\bc^*=\pi_\theta(a_\phi(\bx^*))$.

In this section, we discuss the core questions we aim to answer, followed by a description of the visual abstractions and driving agent considered in our study.

\subsection{Research Questions}
\label{sec:questions}

We aim to build a segmentation dataset $S$ that is cost-effective, yet encodes all relevant information for policy learning. We are interested in the following questions:

\boldquestion{Can selecting specific classes ease policy learning?} A semantic segmentation $\bs$ assigns each pixel to a discrete category $k\in\{1,\dots,K\}$. Knowing whether a pixel belongs to the building class or tree class may provide no additional information to a driving agent, if it knows that the pixel does not belong to the road, vehicle or pedestrian class. We are interested in understanding the impact of the set of categories on the driving task.

\boldquestion{Are semantic representations trained with few images competitive?} In a policy learning setting, the training of the driving agent may be able to automatically compensate for some drop in performance of the segmentation model. We aim to determine if a parsimonious training dataset obtained by reducing the number of training images $n_s$ for the segmentation model can achieve satisfactory performance. 

\boldquestion{Is fine-grained annotation important?} Exploiting coarse annotations such as 2D bounding boxes instead of pixel-accurate segmentation masks can alleviate the key challenge in building segmentation models: annotation costs \cite{Zlateski2018CVPR}. If fine-grained annotation can be avoided, we are interested in how to select $a_\phi$ to exploit coarse annotation during training.

\boldquestion{Are visual abstractions able to reduce the variance which is typically observed when training agents using behavior cloning?} Significant difference in performance of behavior cloning policies is caused as a result of changing the training seed or the sampling of the training data~\cite{Codevilla2019ICCV}. This is problematic in the context of autonomous driving where evaluating an agent is expensive and time-consuming, making it difficult to assess if changes in performance are a result of algorithmic improvements or random training seeds. Since visual priors abstract out certain aspects of the input such as illumination and weather, we are interested in investigating their effect on reducing the variance in policies with different random training seeds.

\subsection{Visual Abstractions}

For our analysis, we consider three visual abstractions based on semantic segmentation.

\boldparagraph{Privileged Segmentation} As an upper bound, the ground-truth semantic labels (available from the simulator) can be used directly as an input to the driving agent. This form of privileged information is useful for ablative analysis.

\boldparagraph{Standard Segmentation} For standard pixel-wise segmentation over all classes, our perception stack is based on a ResNet and Feature Pyramid Network (FPN) backbone \cite{He2016CVPR,Lin2017CVPRb}, with a fully-convolutional segmentation head \cite{Long2015CVPR}. 

\boldparagraph{Hybrid Detection and Segmentation} To exploit coarser annotation, we use a hybrid architecture that distinguishes between stuff and thing classes \cite{Heitz2008ECCV}. The architecture consists of a segmentation model trained on stuff classes annotated with semantic labels (\eg road, lane marking) and a detection model based on Faster-RCNN \cite{Ren2015NIPS} trained on thing classes annotated with bounding boxes (\eg pedestrian, vehicle). The final visual abstraction per pixel is obtained by overlaying the pixels of detected bounding boxes on top of the predicted segmentation, based on a pre-defined class priority. Similar hybrid architectures have been found useful previously in the urban street scene semantic segmentation setting, since detectors typically have different inductive biases than segmentation networks \cite{Saleh2018ECCV}.

\subsection{Driving Agent}

\boldparagraph{Conditional Imitation Learning} In general, behavior cloning involves a supervised learning method which is used to learn a mapping from observations to expert actions. However, sensor input alone is not always sufficient to infer optimal control. For example, at intersections, whether the car should turn or keep straight cannot be inferred from camera images alone without conditioning on the goal. We therefore follow~\cite{Codevilla2018ICRA,Codevilla2019ICCV} and condition on a navigational command. The navigational command represents driver intentions such as the direction to follow at the next intersection. Our agent is a neural network $\pi_\theta$ with parameters $\theta$, which maps a semantic representation $\bs \in \cS$, navigational command $\bn \in \cN$, and measured velocity $v \in \nR^+$ to a control value $\bc \in \cC$:
\begin{equation}
    \pi_\theta : \cS \times \cN \times \nR^+ \rightarrow \cC
    \label{eqn:pi}
\end{equation}

\boldparagraph{Imitation Loss} In order to learn the policy parameters $\theta$, we minimize an imitation loss $\mathcal{L}_{\textnormal{imitation}}$ defined as follows:
\begin{equation}
    \cL_{\textnormal{imitation}} = {|| \bc - \hat{\bc} ||}_1
    \label{eqn:l_p}
\end{equation}
Here, $\hat{\bc}=\pi_\theta(\bs,\bn,v)$ is the control value predicted by our agent and ${||\cdot||}_1$ denotes the $L_1$ norm.

\boldparagraph{Velocity Loss} Recordings of expert drivers have an inherent inertia bias, where most of the samples with low velocity also have low acceleration. It is critical to not overly correlate these since the vehicle would prefer to never start after slowing down. As demonstrated in \cite{Codevilla2019ICCV}, predicting the current vehicle velocity as auxiliary task can alleviate this issue. We thus also use a velocity prediction loss:
\begin{equation}
    \cL_{\textnormal{velocity}} = {|| v - \hat{v} ||}_1
    \label{eqn:l_v}
\end{equation}

\begin{figure}[t]
	\centering
	\includegraphics[width=\columnwidth]{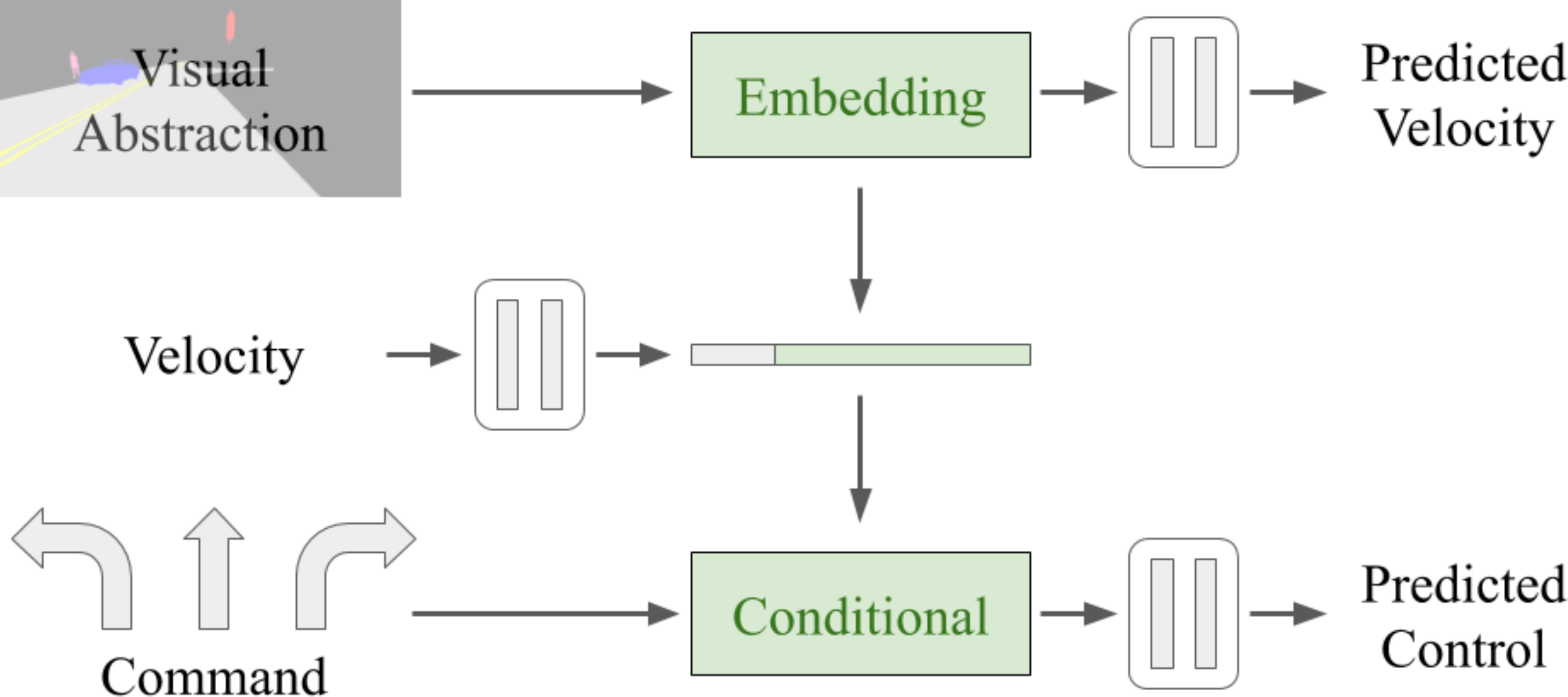}
	\vspace{0.05cm}
	\caption[]{\textbf{Driving agent architecture.} Given a segmentation-based visual abstraction, current vehicle velocity, and discrete navigational command, the CILRS model predicts a control value \cite{Codevilla2019ICCV}.}
	\label{fig:arch}
	\vspace{-0.4cm}
\end{figure}

\boldparagraph{Architecture} The architecture used for our driving agent is based on the CILRS model \cite{Codevilla2019ICCV}, summarized in Fig. \ref{fig:arch}. The visual abstraction is initially processed by an embedding branch, which typically consists of several convolutional layers. We flatten the output of the embedding branch and combine it with the measured vehicle velocity $v$ using fully-connected layers. Since the space of navigational commands is typically discrete for driving, we use a conditional module to select one of several command branches based on the input command. The command branch outputs control values. Additionally, the output of the embedding branch is used for predicting the current vehicle speed, which is compared to the actual vehicle speed in the velocity loss $\mathcal{L}_{\textnormal{velocity}}$ defined in Eq. \ref{eqn:l_v}. The final loss function for training is a weighted sum of the two components, with a scalar weight $\lambda$:
\begin{equation}
    \cL = \cL_{\textnormal{imitiation}} + \lambda \cL_{\textnormal{velocity}}
    \label{eqn:loss}
\end{equation}

\section{Experiments}

In this section, we present a series of experiments on the open-source driving simulator CARLA~\cite{Dosovitskiy2017CORL} to answer the questions raised in Section \ref{sec:questions}. We perform our analysis by training and evaluating driving agents on the NoCrash benchmark of CARLA (version 0.8.4) \cite{Codevilla2019ICCV}.

\subsection{Task}

The CARLA simulation environment consists of two map layouts, called Town 01 \& Town 02, which can be augmented by 14 weather conditions. We use the provided `autopilot' expert mode for data collection. All training and validation data is collected in Town 01 with the four weather conditions specified as the `train' conditions in the NoCrash benchmark. The number of external agents is uniformly sampled from the range $[80, 160]$. Town 02 is reserved for testing.

\boldparagraph{Evaluation} The primary evaluation criterion in CARLA is the percentage of successfully completed episodes during an evaluation phase, referred to as Success Rate (SR). Successful navigation requires driving the vehicle from a starting position to a specified destination within a fixed time limit. Failure can be a result of collision with pedestrians, vehicles or static objects; or inability to reach the destination within the time limit (timeout). The benchmark consists of three levels of traffic density: Empty, Regular and Dense involving 0, 65 and 220 external agents respectively. We perform evaluation in Town 02 for two `test' weather conditions of the NoCrash benchmark that are unseen during training.

\begin{table}[t]
	\centering
	\caption{\textbf{Summary of control datasets.} 
	The third column indicates the number of labeled images used for training our detection and segmentation models. The fourth column indicates the approximate cost of annotating these images.}
	\begin{tabular}{l|c|c|c}
		\textbf{Name} & \textbf{Classes} & \textbf{Labeled Images} & \textbf{Cost (Hours)} \\
		\hline
		Standard-Large-14 & 14 & 6400 & 7500 \\
		Standard-Large & 6 & 6400 & 3200 \\
		Standard & 6 & 1600 & 800 \\
		Standard-Small & 6 & 400 & 200\\
		Hybrid & 6 & 1600 & 50\\
	\end{tabular}
	\label{tab:datasets}
	\vspace{-0.4cm}
\end{table}

\begin{figure*}[t]
	\setlength{\tabcolsep}{0pt}
	\centering
	\begin{tabular}{cccc}
		\includegraphics[height=3.95cm]{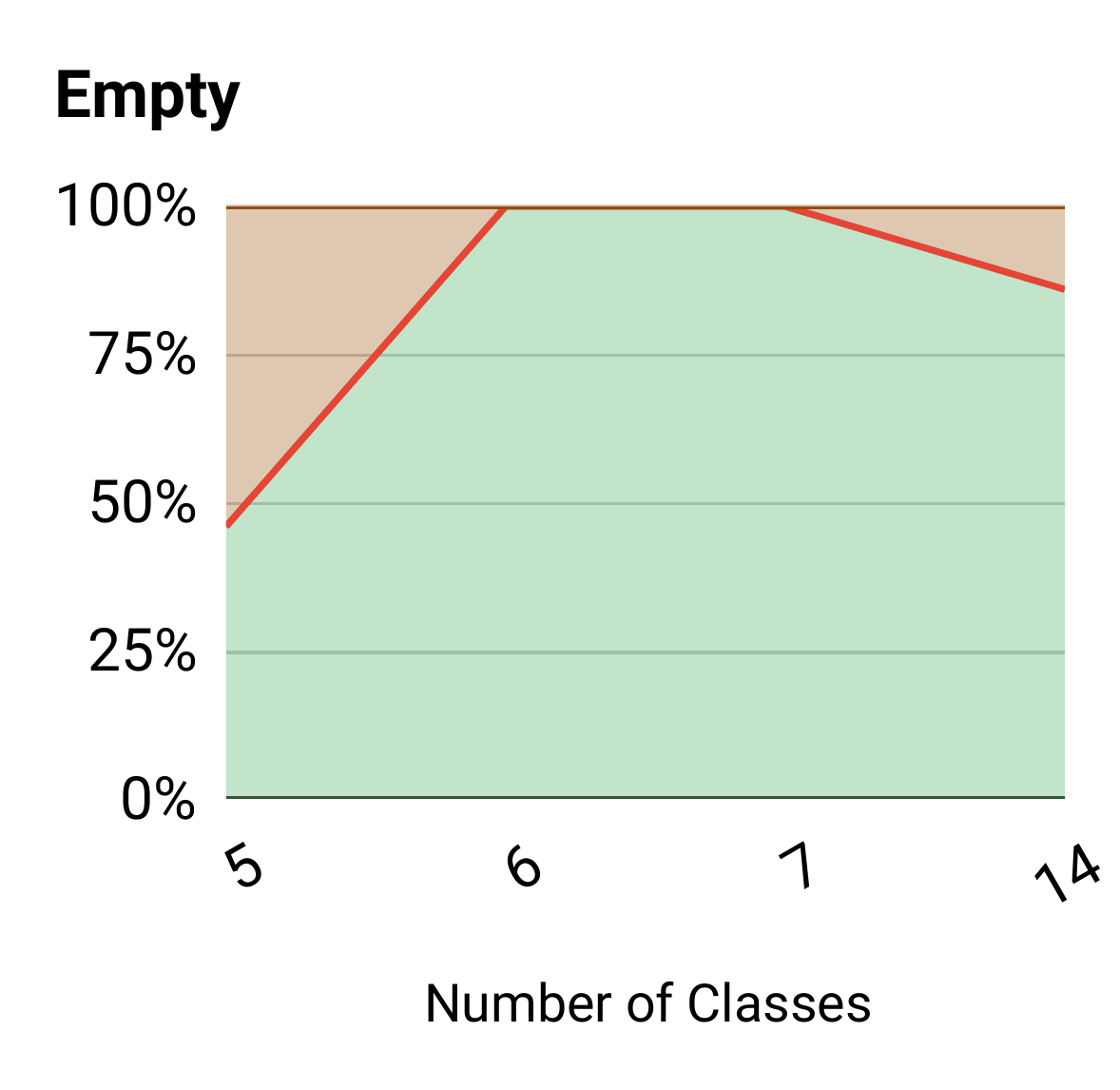} & \includegraphics[height=3.95cm]{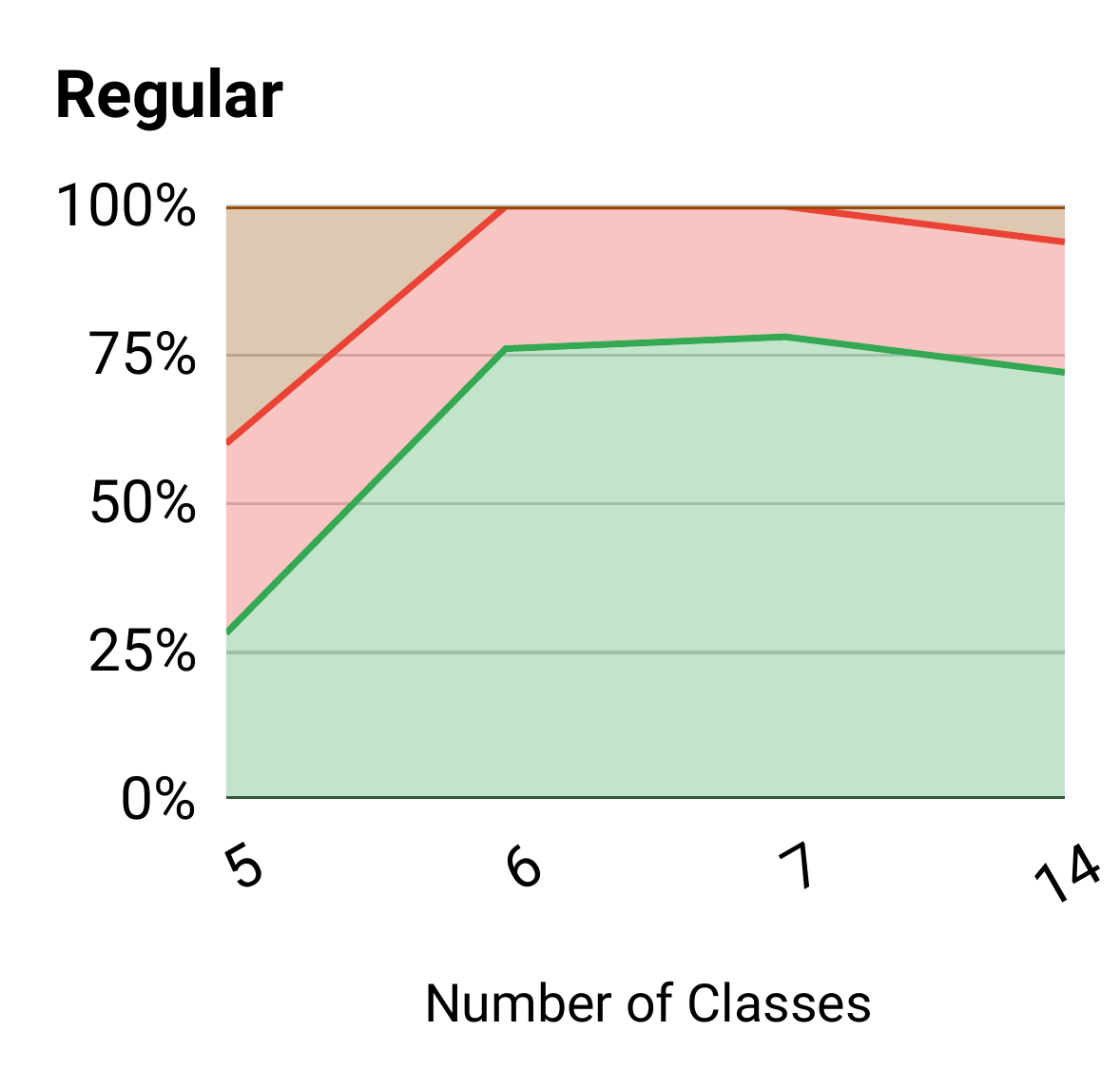} & \includegraphics[height=3.95cm]{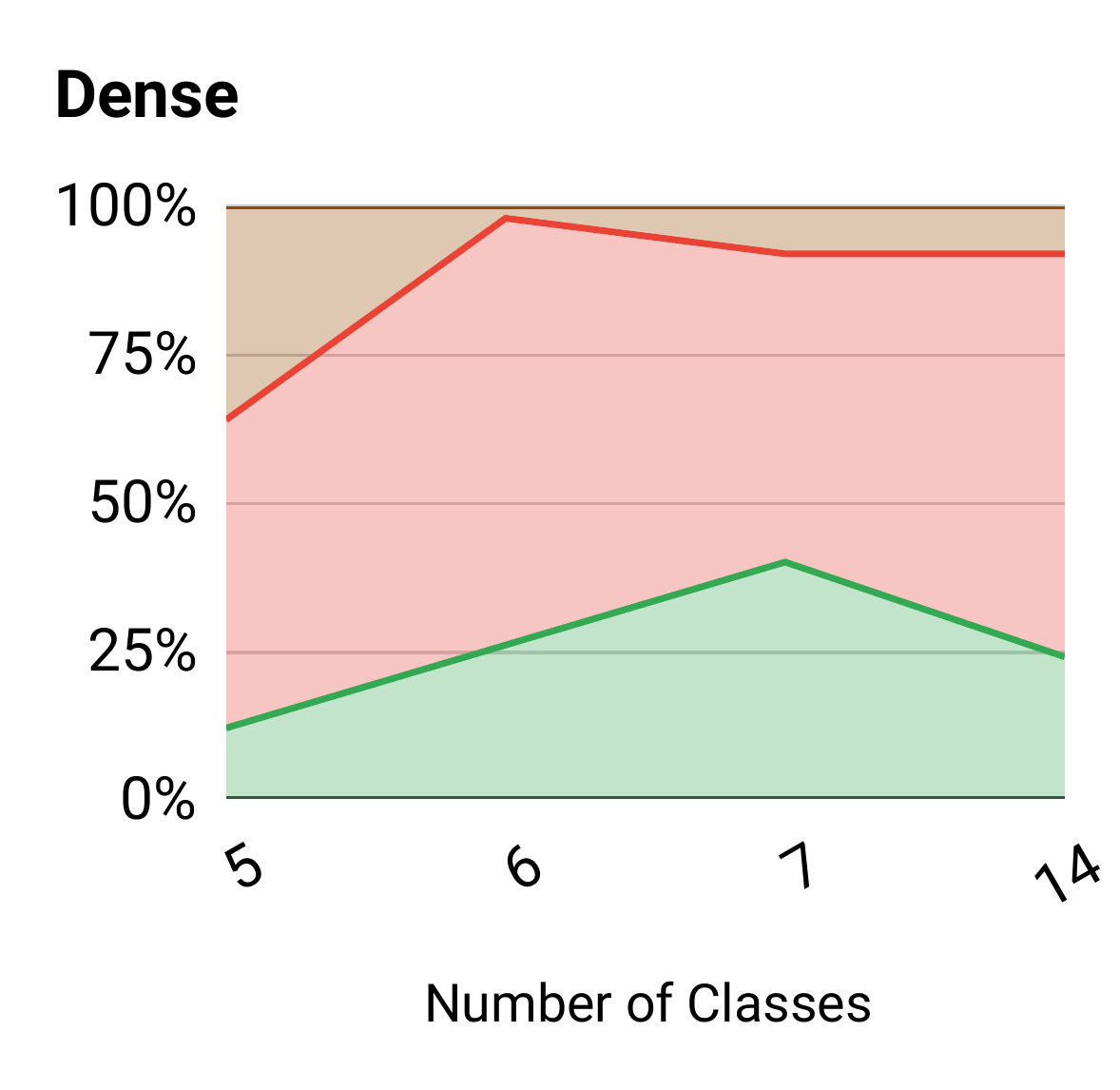} & \includegraphics[height=3.95cm]{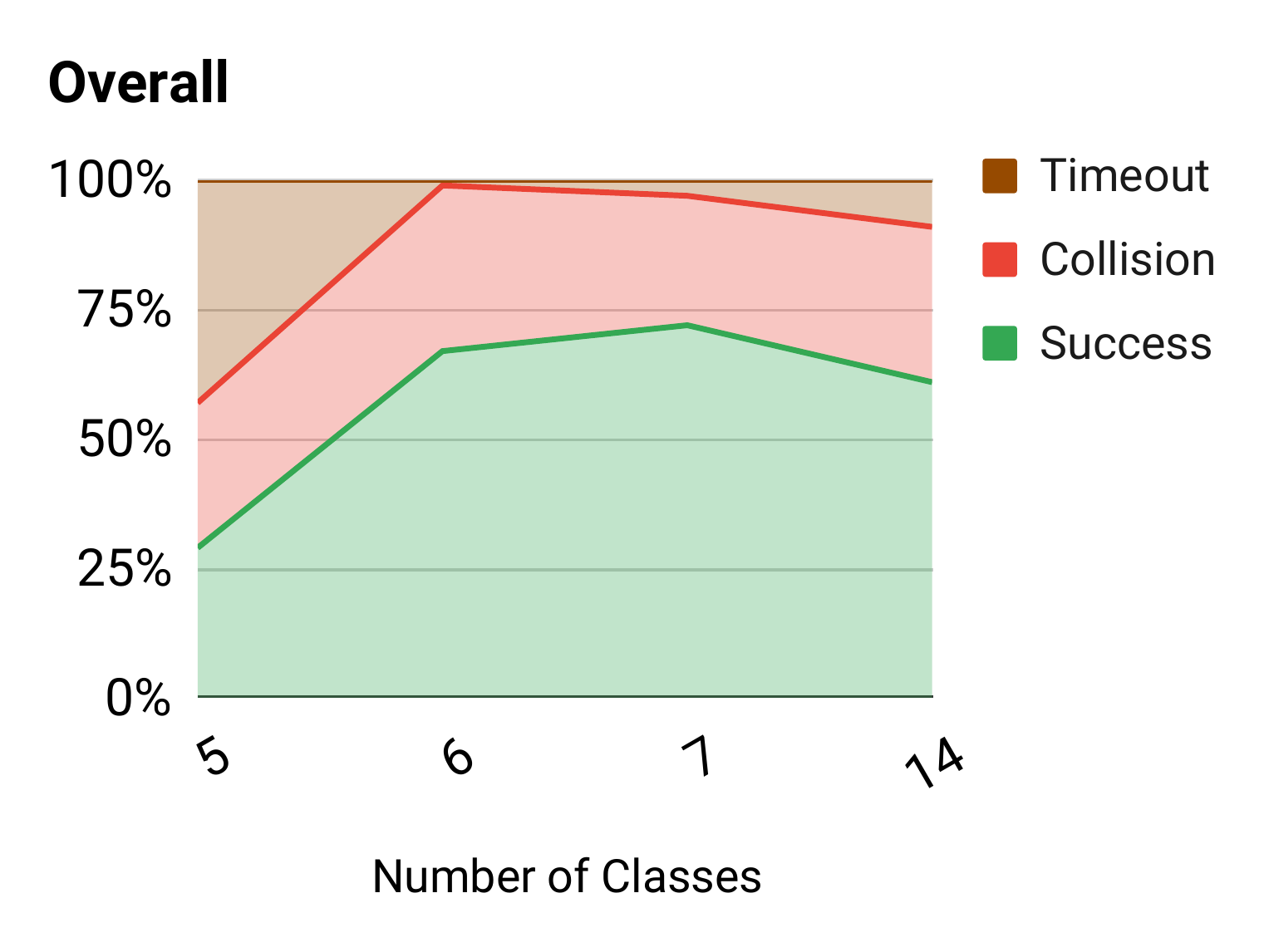} \\
	\end{tabular}
	\caption[]{\textbf{Identifying most relevant classes.} {\color{darkgreen}Success}/{\color{red}collision}/{\color{brown}timeout} percentages on the test environment (Town 02 Test Weather) of the CARLA NoCrash benchmark. For this ablation study, we use ground truth segmentation as inputs to the behavior cloning agent. Reduction from fourteen to seven or six classes leads to a slight increase in success rate, but further reduction to five classes leads to a large number of failures.}
	\label{fig:classes}
\end{figure*}

\begin{figure*}[t]
	\setlength{\tabcolsep}{0pt}
	\centering
	\begin{tabular}{cccc}
		\includegraphics[height=3.25cm]{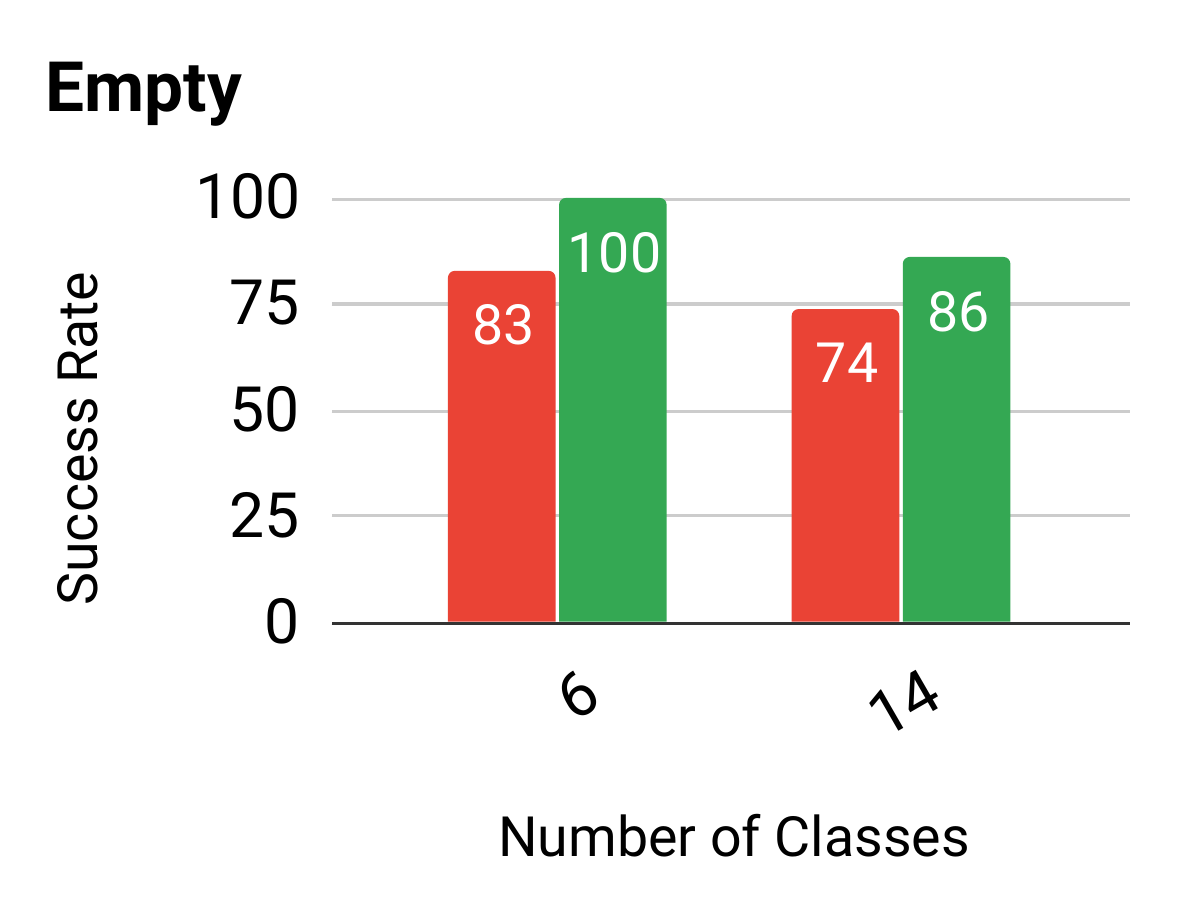} & \includegraphics[height=3.25cm]{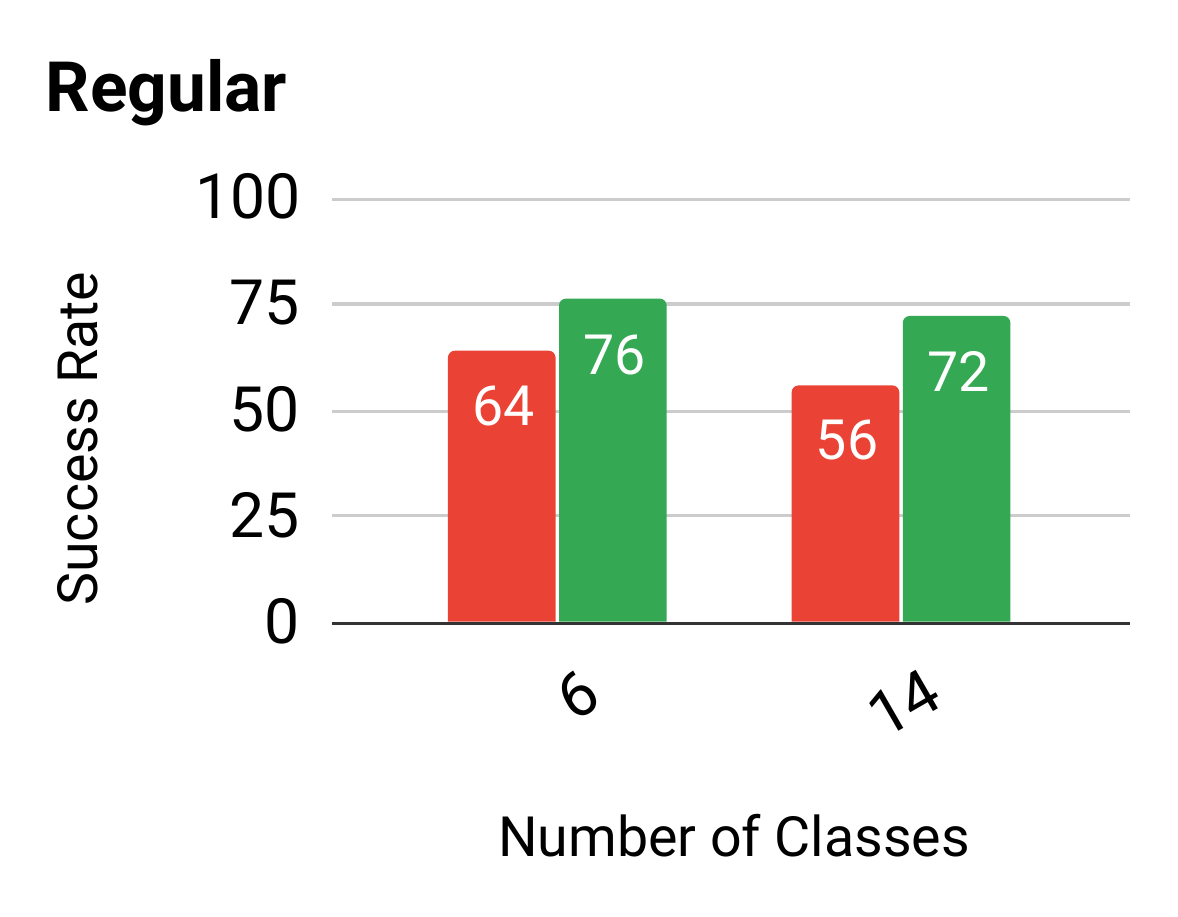} & \includegraphics[height=3.25cm]{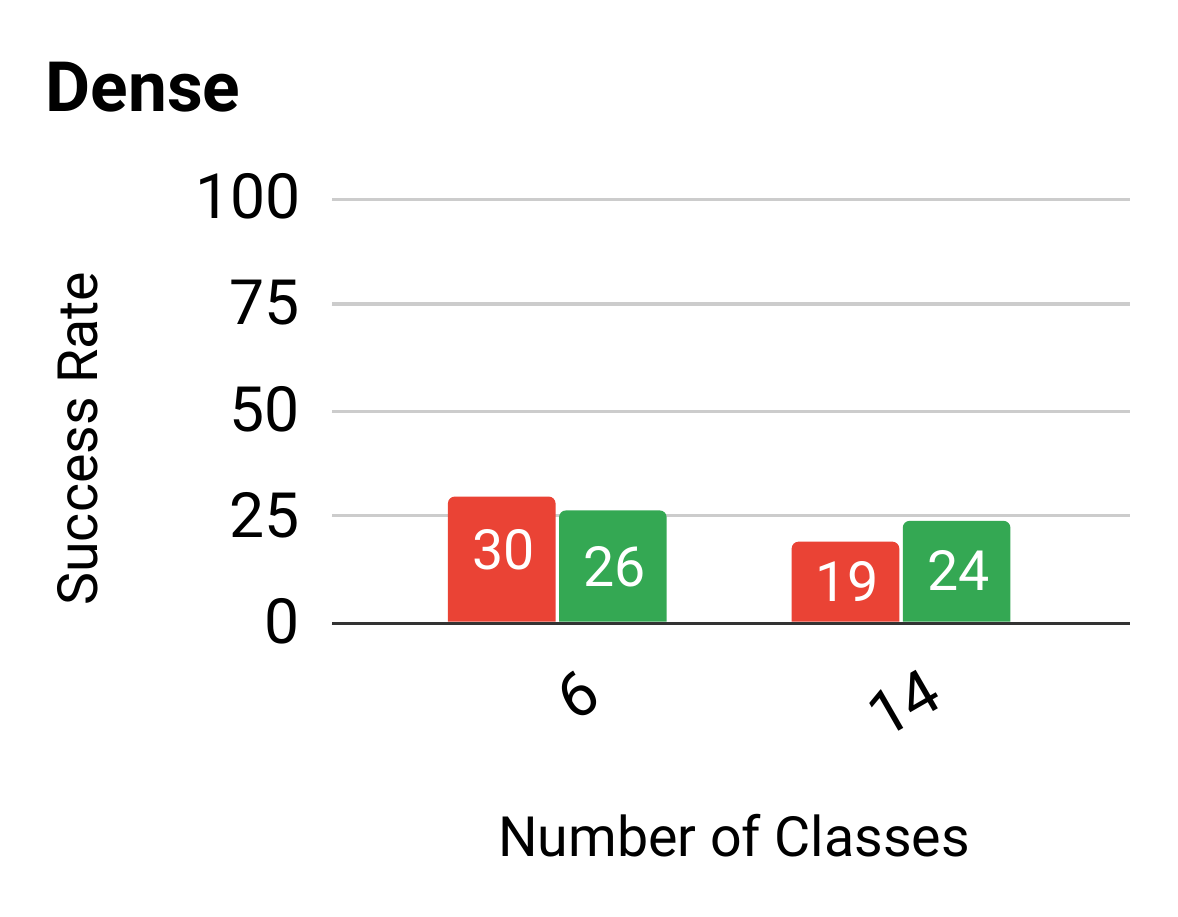} & \includegraphics[height=3.25cm]{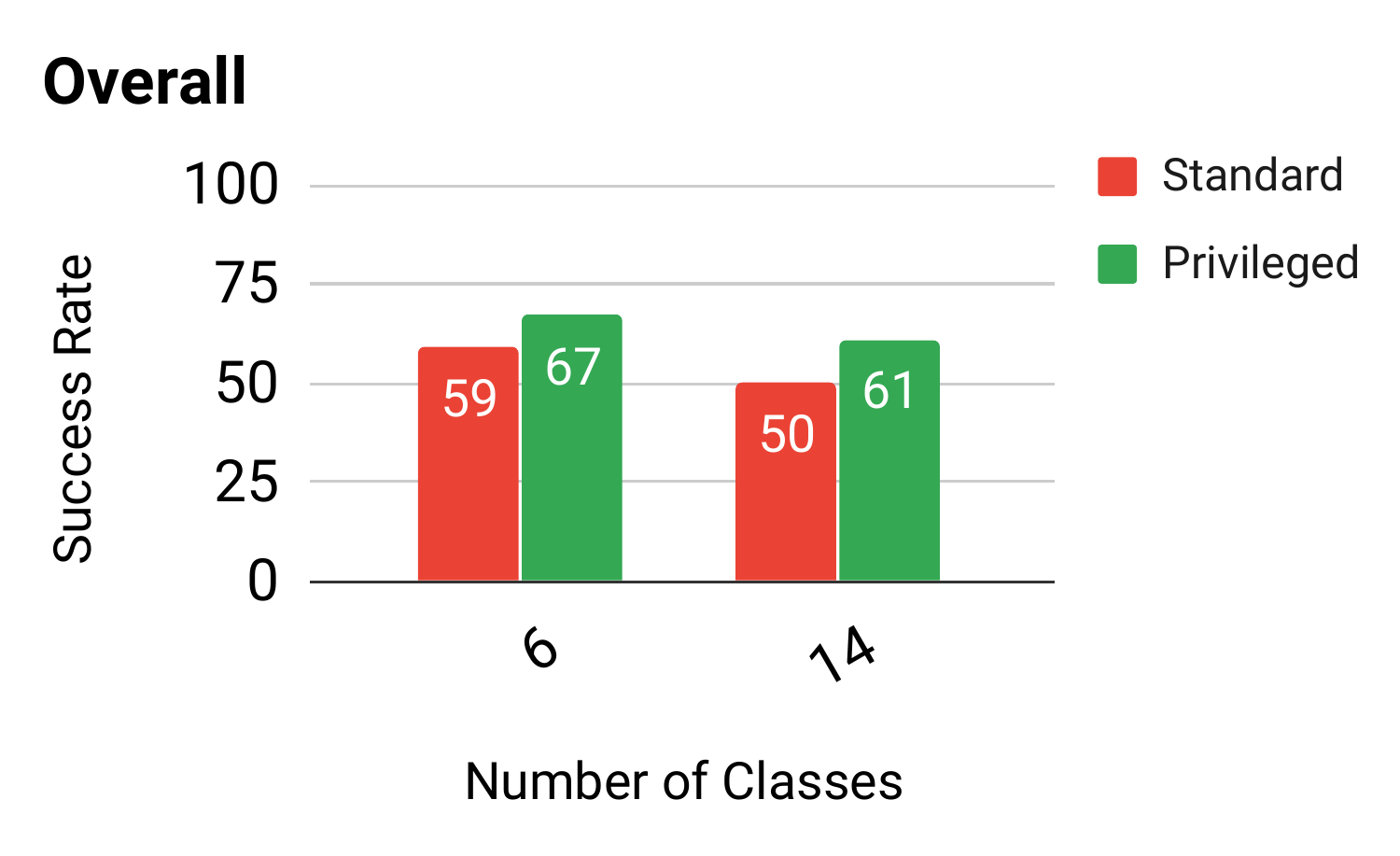} \\
	\end{tabular}
	\caption[]{\textbf{Evaluating the six-class representation.} Success Rate on the test environment (Town 02 Test Weather) of the CARLA NoCrash benchmark. The six-class representation consistently leads to better performance than the fourteen-class representation while simultaneously having lower annotation costs.}
	\label{fig:classes_2}
	\vspace{-0.4cm}
\end{figure*}

\subsection{Datasets}

\boldparagraph{Perception} We collect a training set of 6400 images from Town 01 for training our detection and segmentation models. The images are annotated with 2D semantic labels for 14 classes, and 2D object boxes for 4 of these classes. These annotations are provided by the CARLA simulator.

\boldparagraph{Control} Following \cite{Codevilla2019ICCV}, we collect approximately 10 hours of driving frames and corresponding expert controls for imitation learning using the autopilot of the CARLA simulator. The images are sampled from three cameras facing different directions (left, center and right) on the car at 10 frames per second. We create five variants of this control dataset (summarized in \tabref{tab:datasets}) by transforming the input RGB images to visual abstractions. The Standard-Large-14 dataset is generated using a segmentation network trained to segment the input into fourteen semantic classes: `road', `lane marking', `vehicle', `pedestrian',  `green light', `red light', `sidewalk', `building', `fence', `pole',`vegetation', `wall', `traffic sign' and `other'. The remaining datasets, namely Standard-Large, Standard, Standard-Small, and Hybrid, use a reduced set of six classes: `road', `lane marking', `vehicle', `pedestrian', `green light' and `red light'. While the Standard datasets are generated using a segmentation network, the Hybrid dataset is generated using the combination of a 2-class segmentation model and 4-class detection model.

The study of \cite{Zlateski2018CVPR} provides approximate labeling costs based on a labeling error threshold, for a dataset of similar visual detail as ours. The annotation time reported for the Standard abstraction (fine labeling) is around 300 seconds per image and per class. Our Hybrid visual abstraction roughly corresponds to a 32-pixel labeling error, which requires approximately 20 seconds per image and per class. Based on these statistics, we include the estimated annotation time for each visual abstraction in \tabref{tab:datasets}.

In addition, we also collect a Privileged dataset comprising the ground truth semantic segmentation for the input, which we use for ablation studies involving privileged agents.

\subsection{Implementation Details}
Our perception models are based on a ResNet-50 FPN backbone pre-trained on the MS-COCO dataset \cite{Lin2014ECCV}. We finetune this model for 3k iterations on the perception dataset with the  hyper-parameters set to the default of the Detectron2\footnote{\url{https://github.com/facebookresearch/detectron2}} detection and segmentation algorithms.

The driving agents use a ResNet-18 model in the `embedding' branch (see Fig. \ref{fig:method}). We process the velocity input with two fully-connected layers of 128 units each, which is combined with the ResNet output in another fully-connected layer of 512 units. The velocity prediction branch and each command branch encompass two fully-connected layers of 256 units. We train each model from scratch for 200k iterations using the default training hyper-parameters of the COiLTRAiNE\footnote{\url{https://github.com/felipecode/coiltraine}} framework. This is the standardized repository used to train imitation learning agents on CARLA~\cite{Codevilla2019ICCV,Zhao2019ARXIV}, which currently supports CARLA version 0.8.4.

\begin{figure*}[t]
	\setlength{\tabcolsep}{0pt}
	\centering
	\begin{tabular}{cccc}
		\includegraphics[height=3.95cm]{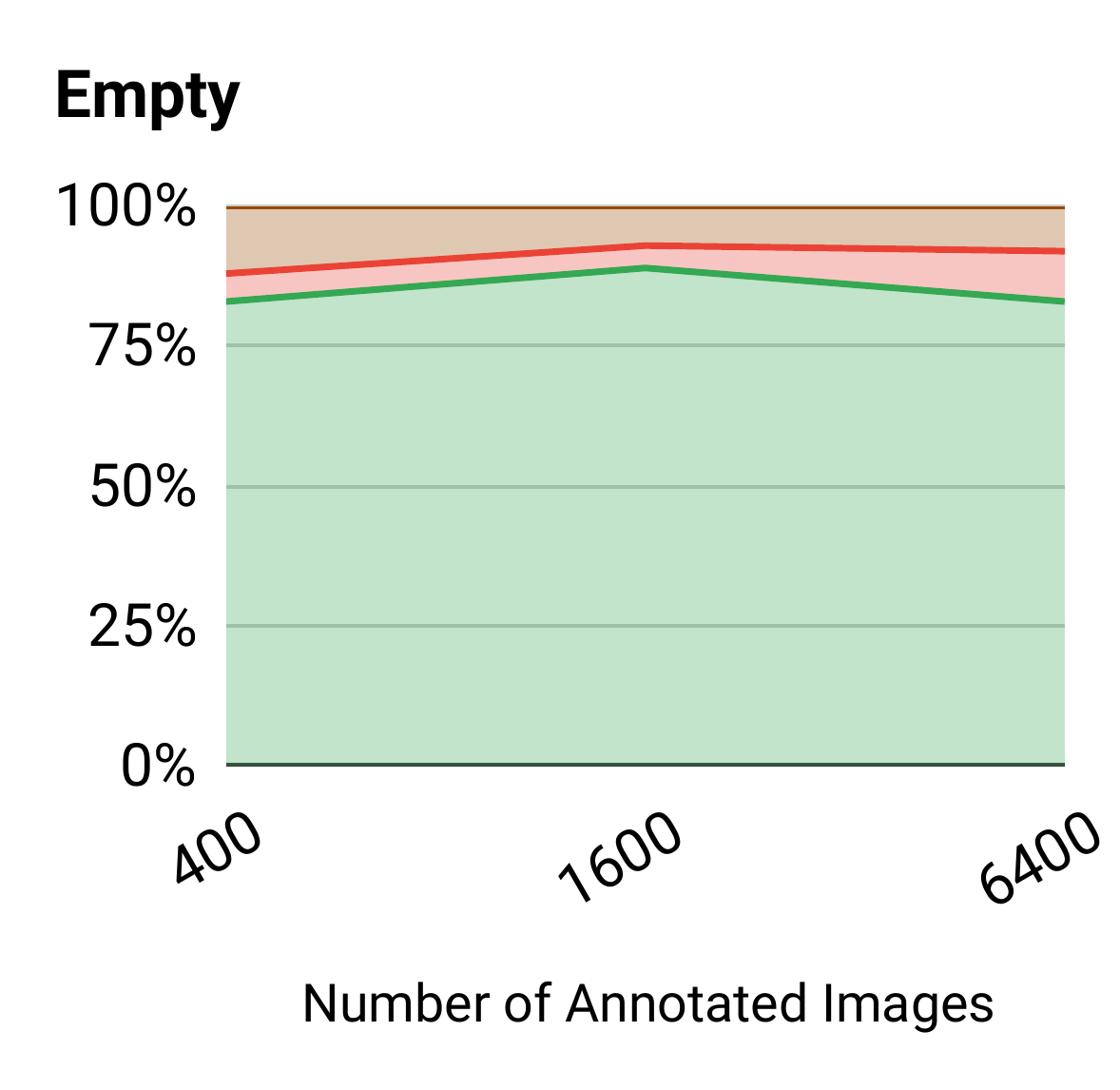} & \includegraphics[height=3.95cm]{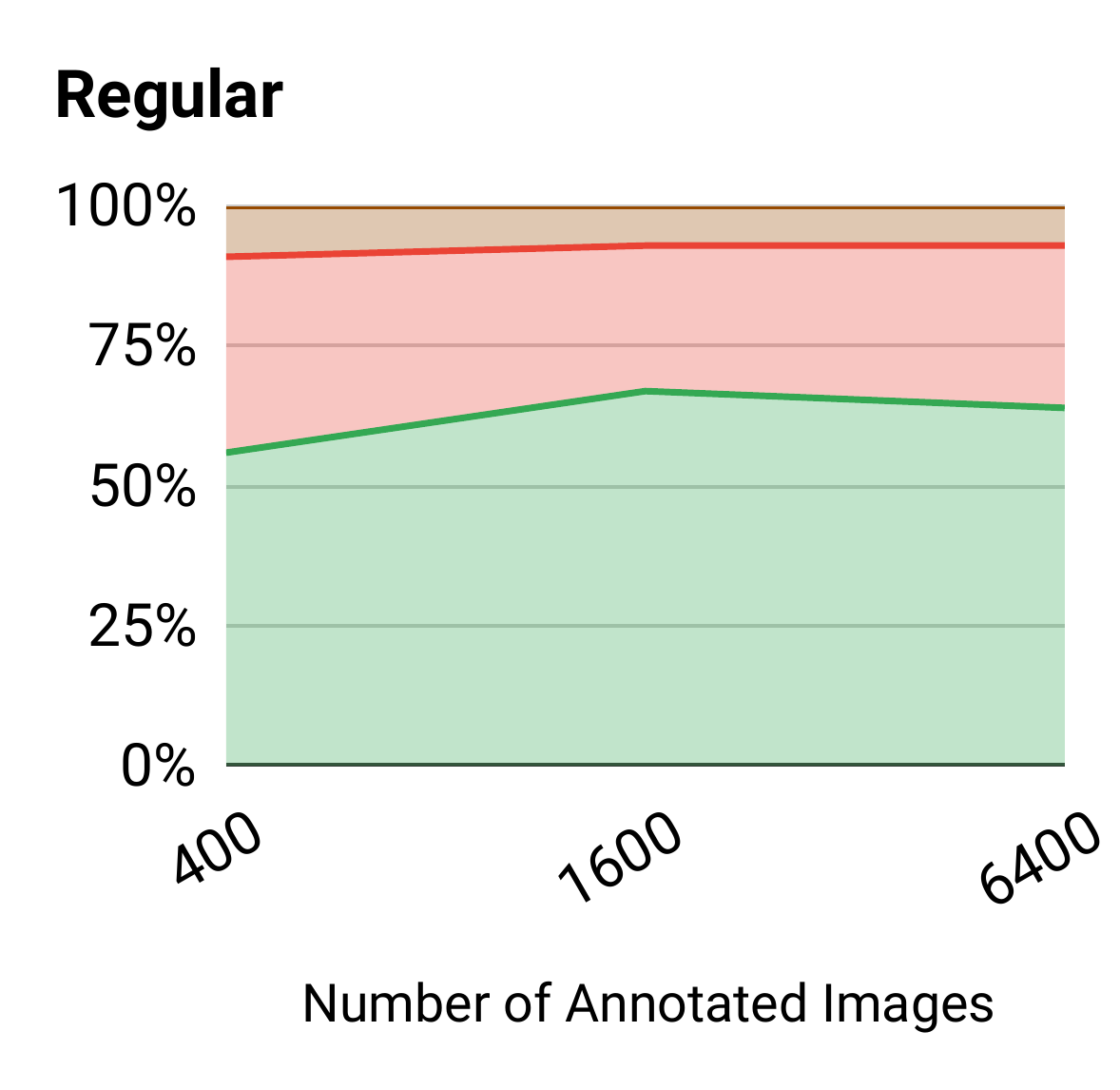} & \includegraphics[height=3.95cm]{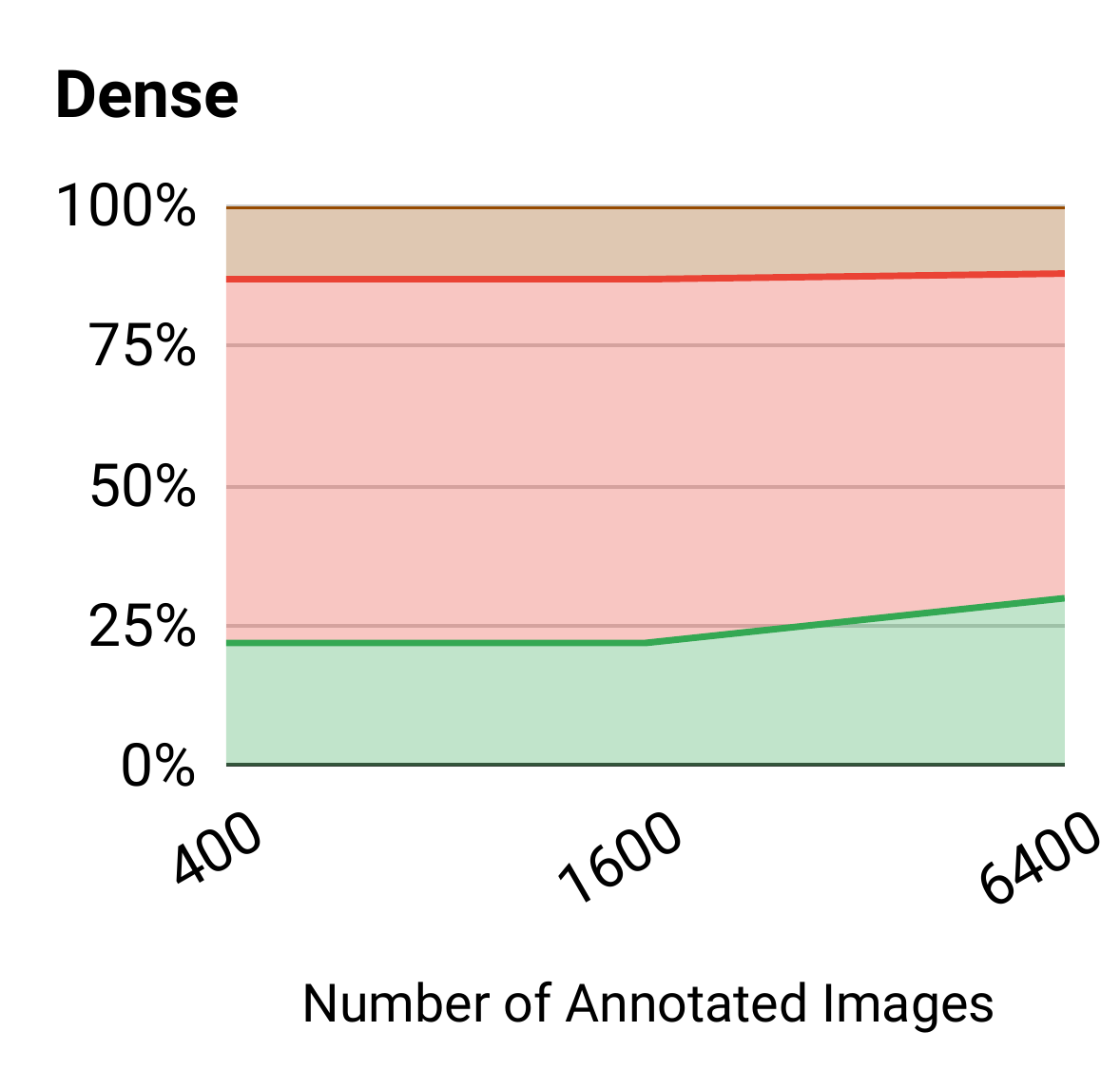} & \includegraphics[height=3.95cm]{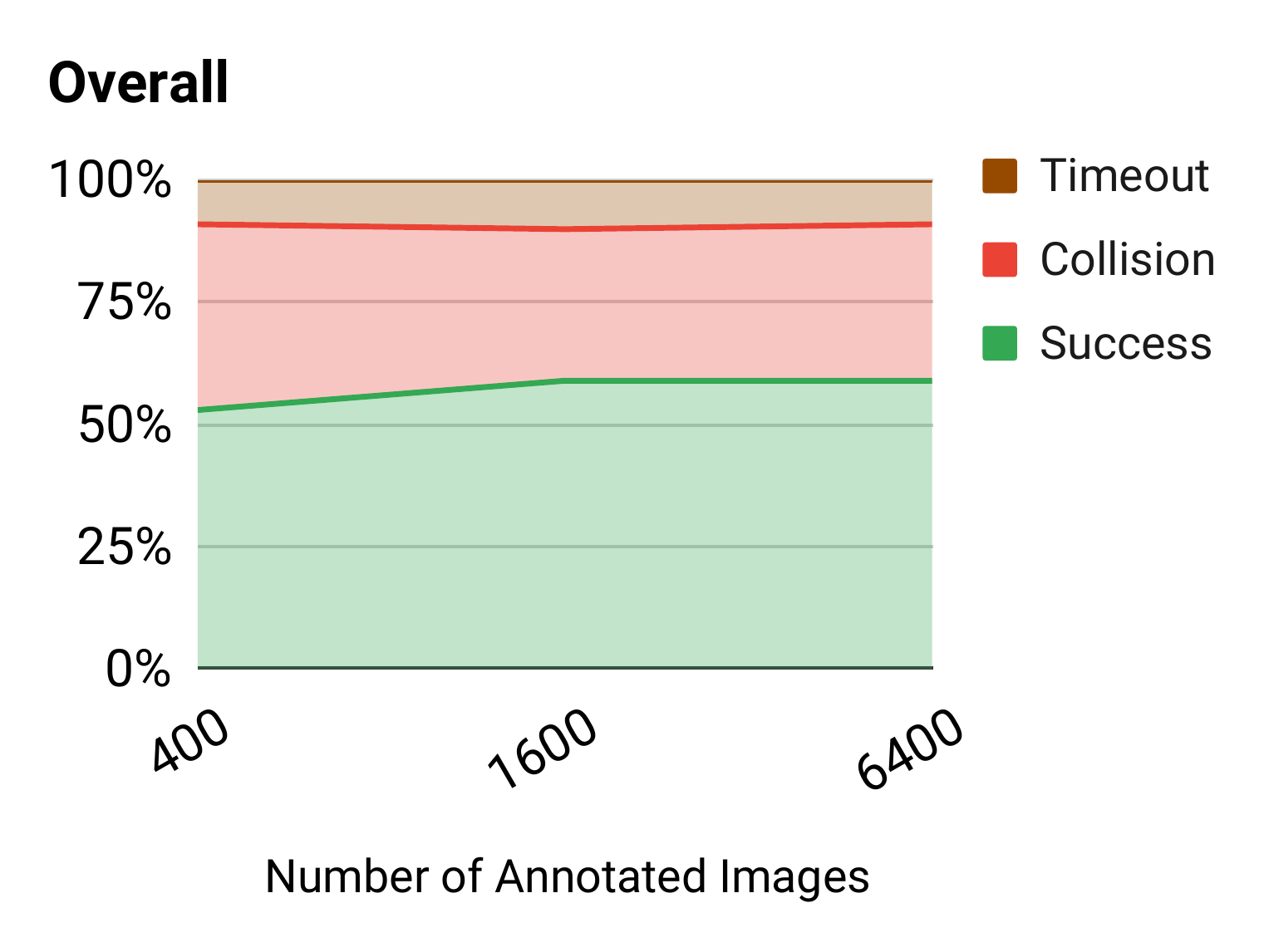} \\
	\end{tabular}
	\caption[]{\textbf{Comparing visual abstractions as annotation quantity is reduced.} {\color{darkgreen}Success}/{\color{red}collision}/{\color{brown}timeout} percentages on the test environment (Town 02 Test Weather). Mean over 5 random training seeds. Performance remains consistent with 6400 or 1600 annotated images, with a slight drop as the training dataset for the visual abstraction is reduced to 400 images.}
	\label{fig:reduce}
	\vspace{-0.4cm}
\end{figure*}

\subsection{Results}

\boldparagraph{Identifying Most Relevant Classes} In our first experiment, our goal is to analyze the impact of training policies with a reduced set of annotated classes. For this, we train and evaluate agents using the Privileged dataset. As a baseline, we use a semantic representation consisting of all fourteen classes in the dataset. We then evaluate a reduced subset of seven classes that we hypothesize to be the most relevant: `road', `sidewalk', `lane marking', `vehicle', `pedestrian', `green light' and `red light'. Next, we evaluate representations that consist of six and five classes, by excluding `sidewalk' and then `lane marking' from the seven-class representation. For the five-class representation, we re-label `lane marking' as `road'. The driving performance for these representations is summarized in Fig. \ref{fig:classes}. We show the percentage of evaluation episodes in the test environment where the agent succeeded, collided with an obstacle or timed out.

We note from our results that perfect segmentation accuracy does not mean perfect overall perception. The fourteen-class model does not achieve perfect driving in any of the three traffic conditions. Even with perfect perception, limitations of using behavior cloning methods such as covariate shift, where the states encountered at train time differ from those encountered at test time, can lead to non-optimal driving behavior. Further, the higher relative dimensionality when using fourteen classes, which includes fine details of classes such as fences, buildings, and vegetation, makes it harder for the agent to identify the right features important for generalization. This is reflected by the fact that the seven-class representation outperforms the agent based on fourteen classes in all three traffic conditions. We empirically observe that the fourteen-class agent is more conservative in its driving style, and more susceptible to timeouts.

The six-class representation that excludes sidewalk segmentation achieves similar performance to seven classes in empty and regular traffic. We therefore additionally compare the six-class and fourteen-class representations using inferred visual abstractions without privileged information, in order to analyze if the same trends observed in Fig. \ref{fig:classes} hold. Specifically, we compare the Standard-Large-14 and Standard-Large datasets as described in Table \ref{tab:datasets}. These datasets are generated using fourteen-class and six-class segmentation networks respectively. The success rates of these trained models are shown in Fig. \ref{fig:classes_2}. Additionally, we show the performance of the corresponding six-class and fourteen-class Privileged agents for reference. We observe that the six-class representation consistently maintains or improves upon the performance of agents trained using all fourteen classes. The six-class approach helps to reduce annotation costs by removing the requirement of assigning labels to several classes such as poles and vegetation, which can be time-consuming due to thin structures with a lot of fine detail. 

Interestingly, we observe from Fig. \ref{fig:classes} that using only five classes leads to a significant reduction in performance, with the overall success rate dropping from 67\% to 29\%. This drastic change indicates that the lane marking class is of very high importance for learning driving policies, and the task becomes hard to solve without this class even with perfect segmentation accuracy on all other classes. Based on the consistent performance of the six-class visual abstraction in both Fig. \ref{fig:classes} and Fig. \ref{fig:classes_2}, we choose this representation to perform a more detailed analysis of trade-offs related to labeling quality and quantity.

\boldparagraph{Number of Annotated Images} In our second experiment, we study the impact of reducing annotation quantity by training agents using the Standard-Large, Standard, and Standard-Small datasets from Table \ref{tab:datasets}. Reducing from Standard-Large to Standard-Small, each dataset has 4 times less samples (and therefore 4 times less labeling cost) than the preceding one. Our results, presented as the mean success, collision and timeout percentages over 5 different training seeds for the behavior cloning agent, are summarized in Fig. \ref{fig:reduce}.

We observe no significant differences in overall downstream driving task performance between the agents trained on 6400 or 1600 samples, and a slight drop when using 400 images. Taking a closer look at the driving performance, we observe that the number of collisions in dense traffic is slightly lower for 6400 samples, but success rate is also slightly decreased on empty conditions. This shows that for our task, when focusing on only the most salient classes, a few hundred images are sufficient to obtain robust driving policies. In contrast, prior work that exploits semantic information on CARLA uses fine-grained annotation for several hours of driving data (up to millions of images) \cite{Zhao2019ARXIV}.

From Fig. \ref{fig:reduce}, we clearly observe a saturation in overall performance beyond 1600 training samples. We therefore study the impact of granularity of annotation in more detail while fixing the dataset size to 1600 samples.

\begin{figure}[t]
	\centering
	\includegraphics[width=0.9\columnwidth]{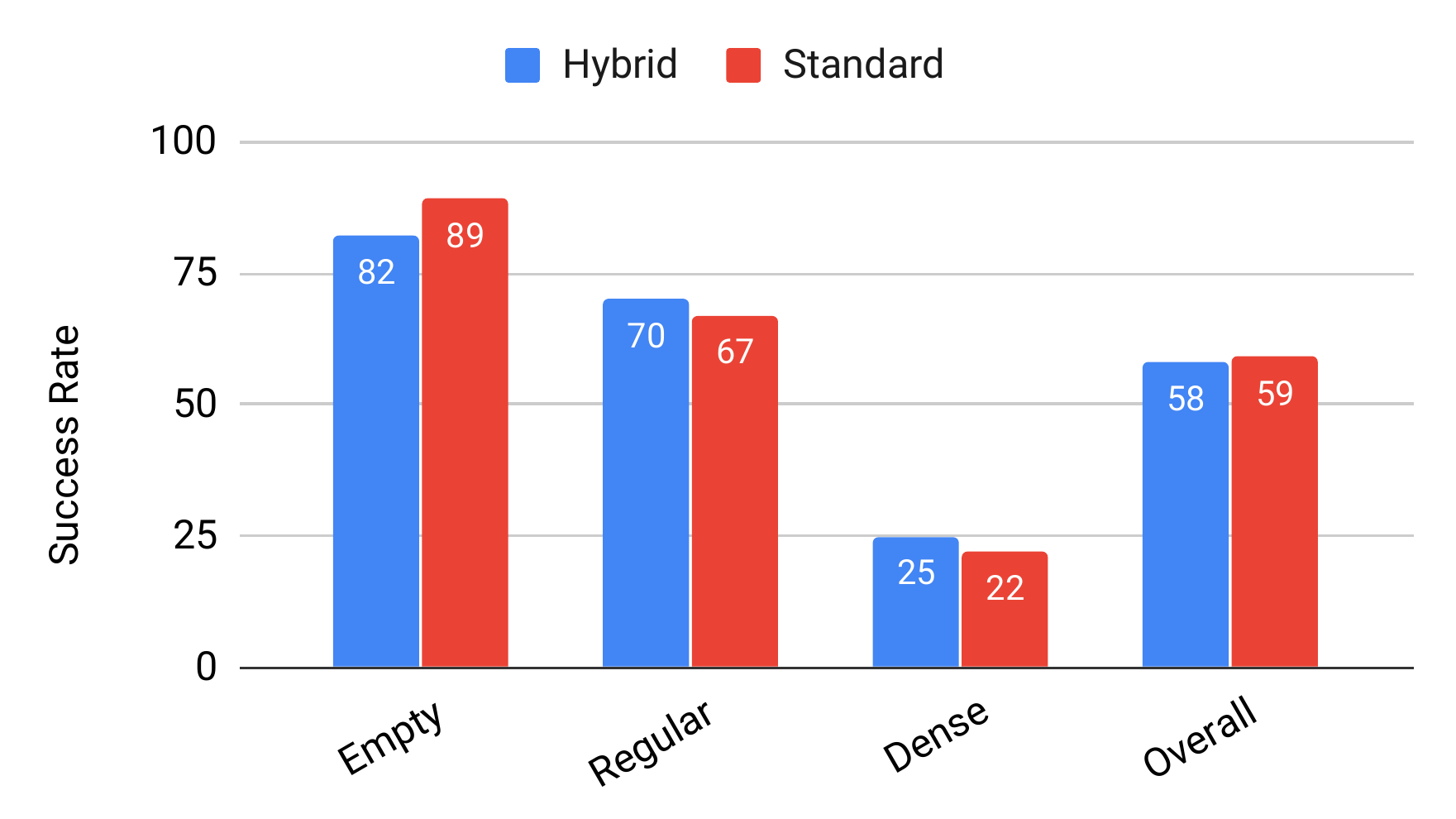}
	\caption[]{\textbf{Evaluating the Hybrid visual abstraction.} Success rate on the test environment (Town 02 Test Weather) as the quality of annotation is reduced. Mean over 5 random training seeds. Overall, the performance of the Hybrid abstraction matches Standard segmentation despite having a reduction in annotation costs of several orders of magnitude.}
	\label{fig:coarse}
	\vspace{-0.4cm}
\end{figure}

\begin{table}[t]
    \setlength{\tabcolsep}{2pt}
	\centering
	\caption{\textbf{Variance between random training seeds.} Percentage Success Rate on Town 02 Test Weather for five training seeds on Empty (E), Regular (R), and Dense (D) conditions, as well as the average Overall (O) success rate. Max and min values indicated in \textbf{bold}. Our approach significantly reduces the standard deviation and coefficient of variation (CV).}
	\label{tab:variance}
	\begin{tabular}{c|c c c c c| c c c}
		\textbf{Task} & \textbf{Seed 1} & \textbf{Seed 2} & \textbf{Seed 3} & \textbf{Seed 4} & \textbf{Seed 5} & \textbf{Mean $\uparrow$} & \textbf{Std $\downarrow$} & \textbf{CV $\downarrow$}\\
		\hline
		\multicolumn{9}{c}{CILRS~\cite{Codevilla2019ICCV}} \\
		\hline
		E  & \textbf{26} & 44 & 42 & \textbf{48} & 46 & 41.20 & 8.79 & 0.21\\
		R  & \textbf{24} & 26 & 30 & 32 & \textbf{40} & 30.40 & 6.23 & 0.20 \\
		D  & \textbf{0} & 2 & 4 & 4 & \textbf{18} & 5.60 & 7.13 & 1.27 \\
		\hline
		O & \textbf{17} & 24 & 25 & 28 & \textbf{34} & 25.60 & 6.18 & 0.24 \\
		\hline
		\multicolumn{9}{c}{Hybrid} \\
		\hline
		E  & \textbf{76} & 80 & 82 & 78 & \textbf{90} & 81.20 & 5.40 & 0.06 \\
		R  & \textbf{64} & 68 & 72 & 72 & \textbf{72} & 69.60 & 3.57 & 0.05 \\
		D  & {28} & 22 & \textbf{18} & \textbf{34} & 22 & 24.80 & 6.26 & 0.25 \\
		\hline
		O & \textbf{55} & 56 & 57 & 61 & \textbf{61} & 58.00 & 2.82 & 0.04 \\
	\end{tabular}
	\vspace{-0.4cm}
\end{table}

\boldparagraph{Coarse Annotation} In our third experiment, we analyze the impact of using the Hybrid visual abstraction that utilizes coarse supervision during training, requiring only an estimated 50 hours to annotate. We present a comparison of the Hybrid and Standard visual abstractions in Fig. \ref{fig:coarse}. The results are presented as the mean driving success rate of five training seeds. We find the Hybrid abstraction to improve performance on tasks involving external dynamic agents, \ie, regular and dense settings. Since objects are input as rectangles, without additional noise introduced by a standard segmentation network, we hypothesize that Hybrid abstractions are able to simplify policy learning (see Fig. \ref{fig:teaser}). Overall, the performance of the Hybrid agent is on par with that of the Standard agent despite having approximately 15 times lower annotation costs (see Table \ref{tab:datasets}).

\boldparagraph{Variance Between Training Runs} In our fourth experiment, we investigate the impact of semantic representations on the variance between results for different training runs. In order to conduct a fair comparison, we use the same raw dataset for training all the models in this study. This ensures that there is no variance caused by the training data distribution. Similar to \cite{Codevilla2019ICCV}, the raw training data was collected by the standard CARLA data-collector framework\footnote{\url{https://github.com/carla-simulator/data-collector}}. 

We compare our approach to CILRS~\cite{Codevilla2019ICCV}, which uses the weights of a network pre-trained on ImageNet~\cite{Russakovsky2015IJCV} to reduce variance due to random initialization of the policy parameters. The only remaining source of variance in training is the random sampling of data mini-batches that occurs during stochastic gradient descent. However, existing studies have still reported high variance in CILRS models between training runs~\cite{Codevilla2019ICCV}. For our approach, we choose the agent trained with the Hybrid abstraction. The results of five different training seeds along with the mean, standard deviation and coefficient of variation (standard deviation normalized by the mean) for each method are shown in Table \ref{tab:variance}.

We observe that for CILRS, the best training seed has double the average success rate of the worst training seed, leading to extremely large variance. In particular, on dense traffic conditions, the success rate ranges from $0$ to $18$. This amount of variance is problematic when trying to analyze or compare different approaches. In contrast, there is less variance observed when training with the Hybrid visual abstraction. Specifically, there is a significant reduction in the standard deviation across all traffic conditions, and the coefficient of variation is reduced by an order of magnitude.

We would like to emphasize that the variance reported in Table \ref{tab:variance} comes from \textit{training} with different random seeds. The training seed is the primary cause of variance, in addition to secondary evaluation variance which is caused by the random dynamics in the simulator. The existing practice for state-of-the-art methods on CARLA is to report only the evaluation variance by running multiple evaluations of a single training seed. We argue (given our findings) that for fair comparison, future studies should additionally report results by varying the training seed and providing the mean and standard deviation (as in Table \ref{tab:variance}).

\begin{table}[t]
    \setlength{\tabcolsep}{3.5pt}
	\centering
	\caption{\textbf{Comparison to state-of-the-art.} Percentage Success Rate on Town 02 of the CARLA NoCrash benchmark, presented as mean and standard deviation over three evaluations of the same model for Empty (E), Regular (R), and Dense (D) conditions. * Indicates our rerun of the best model provided by the authors of \cite{Codevilla2019ICCV}. Models trained with visual abstractions obtain state-of-the-art results.}
	\begin{tabular}{c|c c c c c c |c}
		Task & CAL & CILRS* & LaTeS & LSD & Standard & Hybrid & Expert \\
		\hline
		\multicolumn{7}{c}{Train Weather} \\
		\hline
		E  & $36\pm3$ & $65\pm2$ & $92\pm1$ & ${\bf{94\pm1}}$ & ${91\pm2}$ & $87\pm1$ & $96\pm0$\\
		R  & $26\pm2$ & $46\pm2$ & $74\pm2$ & $68\pm2$ & ${77\pm1}$ & $\bf{82\pm1}$ & $91\pm0$ \\
		D  & $9\pm1$ & $20\pm1$ & ${29\pm3}$ & $30\pm4$ & $27\pm7$ & ${\bf{41\pm1}}$ & $41\pm2$\\
		\hline
		\multicolumn{7}{c}{Test Weather} \\
		\hline
		E  & $25\pm3$ & $71\pm2$ & $83\pm1$ & ${\bf{95\pm1}}$ & ${\bf{95\pm1}}$ & $79\pm1$ & $96\pm0$\\
		R  & $14\pm2$ & $59\pm4$ & $68\pm7$ & $65\pm4$ & ${\bf{75\pm6}}$ & $71\pm1$ & $92\pm0$\\
		D  & $10\pm0$ & ${31\pm3}$ & $29\pm2$ & $\bf{32\pm3}$ & $29\pm5$ & $\bf{32\pm5}$ & $45\pm2$\\
	\end{tabular}
	\label{tab:sota}
	\vspace{-0.4cm}
\end{table}

\boldparagraph{Comparison to State-of-the-Art} In our final experiment, we compare our approach to CAL~\cite{Sauer2018CORL}, CILRS~\cite{Codevilla2019ICCV}, LaTeS~\cite{Zhao2019ARXIV} and LSD~\cite{Ohn-Bar2020CVPR}, which is the state-of-the-art driving agent on the NoCrash benchmark with CARLA version 0.8.4. For fair comparison, we report percentage success rate with the mean and standard deviation over three different evaluations of the best model for each approach. We further report the results of the expert autopilot used for training on the CARLA simulator as an upper bound. Our results are summarized in Table \ref{tab:sota}. We would additionally like to mention that LBC~\cite{Chen2019CORL}, which is the state-of-the-art for a different CARLA version (0.9.6) cannot be directly compared to these methods due to the reliance on several different forms of privileged information (such as the 3D position and orientation of all external dynamic agents).

Conditional Affordance Learning (CAL)~\cite{Sauer2018CORL}, which maps an input image to six scalar `affordances' that are used by a hand-designed controller for driving, is unable to achieve satisfactory performance. We rerun CILRS~\cite{Codevilla2019ICCV} using the author-provided best model, and notice that the rerun numbers (reported in Table~\ref{tab:sota}) differ significantly from the CILRS models we trained in our experiments (reported in Table~\ref{tab:variance}) despite using the same author-provided codebase. The authors do not release the specific dataset used for training their best model, which could explain the difference in performance. However, our models significantly outperform both the CILRS models we trained in our experiments (reported in Table~\ref{tab:variance}) and author-provided best model (reported in Table~\ref{tab:sota}) on every evaluation setting of the benchmark.

The authors of LaTeS~\cite{Zhao2019ARXIV} train a teacher network that takes ground truth segmentation masks as inputs and outputs low-level driving controls. A second student network which outputs driving controls by taking only RGB images as inputs is trained with an additional loss enforcing its latent embeddings to match the teacher network. The training of their teacher network requires fine-grained semantic segmentation labels for each sample used to train the driving policy (hundreds of thousands of images). In contrast, the models trained on our Standard and Hybrid datasets require only a few hundred fine or coarsely labeled images respectively, and outperform LaTeS in the majority of the evaluation settings.

LSD~\cite{Ohn-Bar2020CVPR} uses a mixture model trained using demonstrations and further refined by optimizing directly for the driving task in terms of a reward function. In contrast to our approach, this method uses no image-level annotations, but directly optimizing the policy with a reward is challenging outside of simulated environments. While this approach is slightly better at navigating empty conditions, our models outperform it in regular and dense traffic.
\section{Conclusion}

In this work, we take a step towards understanding how to efficiently leverage segmentation-based representations in order to learn robust driving policies in a cost-effective manner. As fine-grained semantic segmentation annotation is costly to obtain, and methods are often developed independently of the final driving task, we systematically quantify the impact of reducing annotation costs on a learned driving policy. Based on our experiments, we find that more detailed annotation does not necessarily improve actual driving performance. We show that with only a few hundred annotated images, that can be labeled in approximately 50 hours, segmentation-based visual abstractions can lead to significant improvements over end-to-end methods, in terms of both performance and variance with respect to different training seeds. Due to the modularity of this approach, its benefits can be extended to alternate policy learning techniques such as reinforcement learning. We believe that our findings will be useful to guide the development of better segmentation datasets and autonomous driving policies in the future.

\vspace{0.2cm}
\noindent
\textbf{Acknowledgements:} This work was supported by the BMBF through the T\"ubingen AI Center (FKZ: 01IS18039B). The authors also thank the International Max Planck Research School for Intelligent Systems (IMPRS-IS) for supporting Kashyap Chitta and the Humboldt Foundation for supporting Eshed Ohn-Bar. 

{
	\bibliographystyle{ieee}
	\bibliography{bibliography_short,bibliography,bibliography_custom}
}

\end{document}